\documentclass{article}
\usepackage{ijcai20}

\usepackage{times}
\usepackage{soul}
\usepackage{url}
\usepackage[hidelinks]{hyperref}
\usepackage[utf8]{inputenc}
\usepackage[small]{caption}
\usepackage{graphicx}
\usepackage{amsmath}
\usepackage{natbib}
\usepackage{amsthm}
\usepackage{booktabs}
\usepackage{algorithm}
\usepackage{algorithmic}
\usepackage{dsfont}
\usepackage{subfig}
\usepackage{makecell}
\usepackage{bm}
\newcommand{\fronorm}[1]{\left\lVert#1\right\rVert_{F}}
\newcommand{\lnorm}[1]{\left\lVert#1\right\rVert_{2}}

\DeclareUnicodeCharacter{2212}{-}

\def \mat [#1]{\mathbf{#1}}
\def \vec#1{\mathbf {#1}}

\title{Sinkhorn-Flow: Predicting Probability Mass Flow in Dynamical Systems Using Optimal Transport}

\author{
Mukul Bhutani$^1$
\and
J. Zico Kolter$^{1,2}$
\affiliations
$^1$Carnegie Mellon University\\
$^2$Bosch Center for AI\\
\emails
\{mbhutani, zkolter\}@cs.cmu.edu
}

\begin{document}

\maketitle

\begin{abstract}
  Predicting how distributions over discrete variables vary over time is a common task in time series forecasting. But whereas most approaches focus on merely predicting the distribution at subsequent time steps, a crucial piece of information in many settings is to determine how this probability mass \emph{flows} between the different elements over time. We propose a new approach to predicting such mass flow over time using optimal transport. Specifically, we propose a generic approach to predicting transport matrices in end-to-end deep learning systems, replacing the standard softmax operation by Sinkhorn iterations. We apply our approach to the task of predicting how communities will evolve over time in social network settings, and show that the approach improves substantially over alternative prediction methods. We specifically highlight results on the task of predicting faction evolution in Ukrainian parliamentary voting.
\end{abstract}

\section{Introduction}

In this paper, we consider the task of predicting how (discrete) distributions evolve over time.  Such problems are ubiquitous in time series forecasting, with several applications including predicting the market share of different products \citep{demand_forecasting}, predicting energy demand by end-usage type \citep{energy_demand}, and predicting the spread of epidemics \citep{epidemic}.  However, the vast majority of these applications focus solely on predicting how the marginal probabilities of interest vary over time.  But in many instances, it is of critical importance how the probability mass \emph{shifts} between the different entities over time.  For example, we focus in this paper on the specific application of predicting how factions in a network evolve over time.  In this setting, it is crucial not just to forecast the marginal distribution of factions over time, but how the constituencies of the factions shift between groups over time; this can be generically expressed as the problem of predicting the mass flow of these distributions over time.

We present a new approach to predicting such mass flow, which we call the Sinkhorn-Flow model.  At a high level, the approach can be applied to virtually any time series forecasting approach, but with the difference that instead of predicting a single probability distribution at time $t$, we directly predict a transport matrix that determines the distributions at time $t$ and $t+1$, plus the transfer of probability mass between the two distributions. Algorithmically, we accomplish this using a technique similar to the Sinkhorn network approach of \citep{mena2018}, though we extend this approach by directly computing backpropagation through the optimal transport map via implicit differentiation, rather than unrolling the Sinkhorn procedure explicitly.  The basic approach is to generically replace the traditional softmax operator as the ''last layer'' of the time series forecasting task with a Sinkhorn iteration that produces a transport map rather than a single distribution.

We apply our approach to the problem of predicting factions or community evolution in networks, for two separate real-world application areas.  In the first application, we predict how the factions in the Ukrainian parliament evolve over time. These factions are manifested in the way specific members of parliament (MPs) vote on a bill. We show that we can better predict the evolution of those factions than other comparable baselines. In the second application, we try to predict how email communication evolves over time in a European research institute.  In both cases, we show that the proposed method improves substantially over several competitive baselines, including methods that attempt to individually model the members of the community rather than modeling the factions at the ''meta'' level. This is followed by a few qualitative results which visually show the prediction of our model. Finally, we discuss doing multi-step predictions for predicting the flow for more than one timestamp in the future.   

In total, the Sinkhorn-Flow approach substantially advances our understanding of how communities or groups evolve in dynamic networks. And more generally, our methods provide a framework for predicting mass flow in any dynamical system. 

\section{Related Works}

The basis of our proposed model comes from the optimal transport literature. However, since in this paper, we show our algorithm being applied to mass flow prediction in dynamic social networks, we also highlight related work in both the optimal transport and community detection literature.

\subsection{Community detection and evolution}

The problem of learning how factions evolve in a dynamic system is one which underlies many similar tasks. Thus, more specific instances of this problem have been studied by various communities.  One common variant includes predicting how communities evolve in a social network. Although the majority of the work in this domain deals with just community detection, a few recent works have looked into the problem of community evolution as well. 

\citet{community_paris} take the route of predicting the whole network at a future time step and then try to detect communities in that predicted graph. Thus, it implicitly performs community evolution prediction. The method involves extensive network-specific featurization like calculating the number of common neighbors between two nodes, strength of interaction, etc.  \citet{community_featurization1} and \citet{community_featurization2} also use a range of structural features like leadership, density, cohesion and group size for detecting community events in the social network. \citet{community_label} take a slightly more direct approach and focus on identifying event labels such as "survive, growth, shrink, dissolve, split". They use community features along with past event labels in an ARIMA model to predict future event labels for communities.  

All of the approaches listed above are very specific to predicting community evolution in networks. Moreover, the heavy use of network specific hand engineered features make their applicability to problems belonging to other domains difficult. 

In contrast to most of these past approaches, our work focuses on predicting the flow (proportion of mass transfer) between factions, which we believe to be a more general way of approaching the problem than predicting the fate of each individual element. 

\subsection{Optimal Transport and Sinkhorn Networks}

Optimal Transport (OT) \citep{villani} has become a widely adopted approach \citep{visualPerm,salimans2018improving,pmlr-v84-genevay18a} after it was shown that one can compute approximate Wasserstein distances extremely quickly using Sinkhorn iterations \citep{cuturi2013}. 

Sinkhorn iterations \citep{sinkhorn1964} have been used in many application areas in machine learning, from considering robustness properties of deep classifiers \citep{wong2019wasserstein} to learning latent permutations.  \citet{adams2011} use Sinkhorn iterations to produce doubly stochastic matrices as a relaxation of permutation matrices. This approach is closely followed by \citet{mena2018} who use Sinkhorn networks to learn latent permutations.

Following such lines of work, we propose to use Sinkhorn networks for the problem of predicting mass flow between different factions. Akin to the works of \citet{mena2018} and \citet{salimans2018improving}, we learn the optimal transport plan via neural nets. 

\section{Methodology}
\label{sec:methodology}

\subsection{Setting}
\label{sec:dynamic_systems}
Consider a time series over distributions over $k$ elements, that is $\vec x_t \in \Delta^k$, where $\Delta^k$ denotes the $k$-dimensional simplex.   The goal of traditional forecasting approaches is to predict $\vec x_{t+1}$ given all $\vec x_1,\ldots, \vec x_t$, plus any exogenous information at time $t$, which we denote $\vec b_t$.

In our setting, however, we are interested in predicting not just the marginals $\vec x_{t+1}$ alone, but the entire transport map $\mat[P]_t$ such that $(\mat[P]_t)_{ij}$ denotes the amount of probability mass that has moved from $(\vec x_t)_i$ to $(\vec x_{t+1})_j$.  This implies that $\vec x_{t+1} = \mat[P]_t^T 1$ and $\vec x_t = \mat[P]_t 1$, but of course the transport matrix $\mat[P]_t$ constitutes additional information over the marginal distributions.  We are particularly interested in cases where the full transport matrix is \emph{known} at training time so that the predictive model can be trained in a supervised fashion.  This is the case, for instance, in our subsequent application to predicting community shifts. We associate a community with each individual at each time; we, therefore, know precisely how mass shifts from one community to another, based upon the shifts of individual members.

\subsection{Differentiable Sinkhorn iterations}

We propose to learn $\vec x_{t+1}$ using Sinkhorn iterations, following the strategy of Sinkhorn networks \citep{mena2018,adams2011}, though with the additional extension of using implicit differentiation to perform the backward pass rather than explicit unrolling of the Sinkhorn operator.

To briefly review, we can write Sinkhorn operator (or Sinkhorn iterations) on any matrix $\mat[M]$ as: 

\begin{align}
    \mat[S^{0}](\mat[M]) &= \text{exp}(-\mat[M]) \nonumber \\
    \mat[S^{i}](\mat[M]) &= \mathcal{N}_c(\mathcal{N}_r(S^{i-1}(\mat[M])))
    \label{eq:sinkhorn}
\end{align}

where, \, $\mathcal{N}_r(\mat[M]) = \mat[M] \oslash (\mat[M] \mathds{1}_{d}\mathds{1}_{d}^{\top})$ normalizes the rows and $\mathcal{N}_c (\mat[M]) = \mat[M] \oslash (\mathds{1}_{d}\mathds{1}_{d}^{\top} \mat[M])$ normalizes columns (here $\oslash$ represents element wise division). \citet{sinkhorn1964} proved that 
\begin{equation*}
    \mat[S] = \mat[S^{\infty}](\mat[M]) = \underset{i \rightarrow \infty}{\text{lim}} S^i(\mat[M])
\label{eq:sinkhorn_limit}
\end{equation*}
belongs to Birkhoff polytope, which is a set of double stochastic matrices. In practice though, instead of going till infinity, the number of iterations $i$ is truncated to a large enough number $l$. 

One practical consideration arises due to the limit in the equation (\ref{eq:sinkhorn_limit}). Modern deep learning frameworks use auto-differentiation to calculate gradients in the backpropagation step \citep{tensorflow2015-whitepaper}. Having an iterative algorithm as in equation (\ref{eq:sinkhorn}) with a large enough number of iterations can make the process of backpropagation computationally expensive and can also cause memory issues. Current implementations of Sinkhorn networks \citep{mena2018, adams2011} do not take any explicit action to prevent this.

We propose to instead use an implicit differentiation approach to directly backpropagate through the solution of the Sinkhorn iteration procedure, without storing any intermediate iterations or unrolling the computation.  That is, we propose a method for directly multiplying by the Jacobian $\frac{\partial S}{\partial \mathbf{M}}$. Unlike traditional applications of the implicit function theorem, however, it is difficult to derive such a procedure directly from the Sinkhorn iterations themselves, due to the fact that the $\mathbf{M}$ matrix is only used at initialization of the Sinkhorn procedure, and isn't included in each iteration.  Thus, to derive an efficient backward procedure, we first consider the optimization formulation of entropy regularized optimal transport \citep{mena2018}, which we write as
\begin{equation}
\label{entropy}
\begin{array}{cl}{\underset{S}{\operatorname{minimize}}} & {\langle \mat[M], \mat[S]\rangle- H(\mat[S])} \\ {\text { subject to }} & {\mat[S]\mathds{1} =\mathds{1}} \\ {} & {\mat[S^{\top}]\mathds{1} =\mathds{1} }\end{array}
\end{equation}
Where $H(S)=-\sum_{i, j} S_{i j} \log S_{i j}$ is the entropy regularization term. Note, usually, equation \ref{entropy} is accompanied by a parameter $\lambda$.
However, to keep the upcoming derivations clean, we absorb it in the matrix $\mat[M]$.

We want to provide a method to differentiate through the solution of the Sinkhorn iteration, i.e, compute the Jacobian $\frac{\delta{S}}{\delta{\mat[M]}}$ or more concretely, to compute the Jacobian-vector left product for use in backpropagation.

To derive this Jacobian, we will imply implicit differentiation to the optimality conditions of this optimization problem. The resulting solution will lead to an algorithm for computing these Jacobians using an iterative method very similar to the Sinkhorn iterations.

The KKT optimality conditions for equation (\ref{entropy}): 
\begin{align*} 
    \mat[M] + \mathds{1} + \log \mat[S]^{\star}+ \bm \alpha^{\star} \mathds{1}^{\top} + \mathds{1} \bm \beta^{\star \top} &=0 \\ 
    \mat[S]^{\star} \mathds{1} &= \mathds{1} \\ 
    \mat[S]^{\star \top} \mathds{1} &= \mathds{1} 
\end{align*}
where $\mat[S]^{\star}$, $\bm \alpha^{\star}$, and $\bm \beta^{\star}$ are optimal primal and dual variables respectively. 

Representing it in a vector form we have:
\begin{align*} 
\vec m + \mathds{1} + \log \vec s^{\star} + (\mathds{1} \otimes \mat[I]) \bm \alpha + (\mat[I] \otimes \mathds{1}) \bm \beta &=0 \\
\left(\mathds{1}^{\top} \otimes \mat[I]\right) \vec s^{\star} &= \mathds{1} \\
\left(\mat[I] \otimes \mathds{1}^{\top}\right) \vec s^{\star} &= \mathds{1} 
\end{align*}
The standard approach for differentiating through an optimization problem is to consider the Jacobian of these optimality conditions, and the backward pass is equivalent to multiplying by the inverse of this Jacobian. The Jacobian of these optimality conditions is given by the matrix:

\begin{equation*}
\begin{split}
\left[\begin{array}{ccc}{\operatorname{diag}\left(1 / \vec s^{\star}\right)} & {(\mathds{1} \otimes \mat[I])} & {(\mat[I] \otimes \mathds{1})} \\ {\left(\mathds{1}^{\top} \otimes \mat[I]\right)} & {0} & {0} \\ {(\mat[I] \otimes 1)} & {0} & {0}\end{array}\right]
\end{split}
\end{equation*}

Then given some Jacobian of the loss with respect to $\vec s^{\star}$, we can compute the Jacobian with respect to $\vec m$ via the linear solve:
\begin{equation*}
\begin{split}
& \left[\begin{array}{c}{\frac{\partial \ell}{\partial \vec m}} \\ {*} \\ {*}\end{array}\right] = \\ &\quad \left[\begin{array}{ccc}{\operatorname{diag}\left(1 / \vec s^{\star}\right)} & {(\mathds{1} \otimes \mat[I])} & {(\mat[I] \otimes \mathds{1})} \\ {\left(\mathds{1}^{\top} \otimes \mat[I]\right)} & {0} & {0} \\ {(\mat[I] \otimes 1)} & {0} & {0}\end{array}\right]^{-1}\left[\begin{array}{c}{\frac{\partial \ell}{\partial \vec s^{\star}}} \\ {0} \\ {0}\end{array}\right]
\end{split}
\end{equation*}
The actual forming of these matrices would be impractical, though the linear system can be simplified substantially. First note that by standard elimination procedures and by converting the resulting Kronecker products to more efficient matrix operations, we can simplify this linear system to be:
\begin{equation}
\frac{\partial \ell}{\partial \mat[M]}= \mat[S]^{\star} \circ\left(\vec a \mathds{1}^{\top}+ \mathds{1} \vec b^{\top}-\frac{\partial \ell}{\partial \mat[S]^{\star}}\right)
\label{eq-graident-equation}
\end{equation}
where, 

\begin{equation*}
\left[\begin{array}{c}{\vec a} \\ {\vec b}\end{array}\right]  = \\ \left[\begin{array}{c c}{ \mat[I] } & {\mat[S]^{\star}} \\ 
{\mat[S]^{\star \top}} & {\mat[I]}\end{array}\right]^{\dagger}\left[\begin{array}{c}{\left(\mat[S]^{\star} \circ \frac{\partial \ell}{\partial \mat[S]^{\star}}\right) \mathds{1}} \\ {\left(\mat[S]^{\star} \circ \frac{\partial \ell}{\partial \mat[S]^{\star}}\right)^{\top} \mathds{1}}\end{array}\right]
\end{equation*}

Note however that this still involves the solution of the linear system involving the matrix:

$$
\left[\begin{array}{cc}{\mathbf{I}} & {\mathbf{S}^{\star}} \\ {\mathbf{S}^{\star \top}} & {\mathbf{I}}\end{array}\right]
$$

This matrix is singular, but the null space (1, -1) corresponds to precisely those terms that are removed when we form the terms $\vec a \mathds{1}^{\top}+ \mathds{1} \vec b^{\top}$ (i.e., we can arbitrarily add a constant to $\vec a$ and subtract it from $\vec b$. However, since by the same criteria its eigenvalues are bounded between [−1, 1], we can use a simple Richardson iteration to find the solution to this linear system, rather than attempt via a direct solution method. Specifically, this leads to the iteration:

\begin{align*} 
    \vec a^{0}, \vec b^{0} &=0 \\ 
    \bar{\vec a}^{k+1} &=\left(\mat[S]^{\star} \circ \frac{\partial \ell}{\partial \mat[S]^{\star}}\right) \mathds{1} - \mat[S]^\star \vec b^{k} \\ 
    \bar{\vec b}^{k+1} &=\left(\mat[S]^{\star} \circ \frac{\partial \ell}{\partial \mat[S]^{\star}}\right)^{\top} \mathds{1} -{\mat[S]^\star}^{\top} \vec a^{k} \\ 
    \vec a^{k+1} &= \bar{\vec a}^{k+1} - \mathds{1}^{\top}\left(\bar{\vec a}^{k+1}-\bar{\vec b}^{k+1}\right) /(2 n) \\
    \vec b^{k+1} &=\bar{\vec b}^{k+1}+ \mathds{1}^{\top}\left(\bar{\vec a}^{k+1}-\bar{\vec b}^{k+1}\right) /(2 n)
\end{align*}

This procedure typically converges in the same number of iterations, or fewer, as the forward Sinkhorn iterations.  Thus, our final approach uses the above iteration to compute $\vec a$ and $\vec b$, and then uses \eqref{eq-graident-equation} to compute the gradient.  Each iteration of this procedure involves only an elementwise Haramard product and a Matrix-vector product with the $S$ matrix, and thus takes $O(n^2)$ computation time, the same complexity as a forward Sinkhorn iteration.

\subsection{Predicting the transport map}
We now consider the task of using these optimal transport predictions to predict mass flow in a dynamical system. Given past information (as described in section \ref{sec:dynamic_systems}), our goal is to predict the full transport map at time $t$, that is
\begin{equation}
    \hat{\mat[P]}_{t} = g(\vec b_t, \mat[x]_{t},..., \mat[x]_{t-m}; \Theta), 
    \label{eq:transport_pred}
\end{equation}
where $\mat[P]_t \mathds{1} = \vec x_t$ and $\mat[P]_t^T \mathds{1} = \vec x_{t+1}$ as mentioned above, and where in practice we truncate the history of the auto-regressive portion of our model so as to only look at the past $m$ time steps (past exogenous information can all be bundled into the $\vec b_t$ term, so explicit lag terms are required there).  To make this prediction, we propose to use the function
\begin{equation}
    \hat{\mat[P]}_{t} = \mathrm{diag}(\vec x_t) S(f(\vec b_t, \mat[x]_{t},..., \mat[x]_{t-m}; \Theta)), 
    \label{eq:transport_pred2}
\end{equation}
where $S$ denotes the Sinkhorn operation mentioned above, and $f$ denotes an arbitrary function (we use simple neural networks in our setting) that outputs a $k \times k$ matrix.  To motivate this precise form, note that the Sinkhorn iteration above produces a doubly stochastic matrix (i.e., the probability transition matrix) rather than a transport map, and to convert this to a transport map (which has $\mat[P]_t \mathds{1} = \vec x_t$) we need to pre-multiply by $\mathrm{diag}(\mat[x]_t)$.

Given this predicted transport map, we could define several possible losses between it and the actual ground truth transport map $\mat[P]_t$, i.e., via KL divergence terms or something similar.  However, in practice it appears sufficient to simply penalize a squared distance term between the predicted and actual distribution.  We did note, however, that it was helpful to interpolate between a loss that penalizes the entire transport map and one that penalizes just the one-step marginal predictions.  That is, our final optimization problem is given by

\begin{equation}
    \begin{split}
    \underset{\Theta}{\text{minimize}} \;\; \sum_{t=1}^T \biggl( (1 - \lambda)\fronorm{\mat[P]_{t} - \hat{\mat[P]}(\Theta)_{t}}^{2} \\
     +  \lambda \lnorm {\vec x_{t+1} -    \hat{\mat[P]}_{t}^{\top} (\Theta) \mathds{1} }^2 \biggr),
    \label{eq:objective}
    \end{split}
\end{equation}

where $\fronorm{.}$ and $\lnorm{.}$ refer to Frobenious and $\textit{l}_{2}$ norms respectively and $\lambda$ is a hyperparameter.

\section{Experiments}

To show the efficacy of our algorithm, Sinkhorn-Flow, we take the problem of predicting evolution of factions in social networks. We perform the experiments using two real world datasets and compare our algorithm with competitive baselines. 

\subsection{Ukrainian Parliamentary Factions Prediction}

The first problem is to predict how factions evolve in the an elected parliament. Ukraine went through a political troublesome time due to the Euromaidan crisis  and the Russian invasion of Crimea in 2013-2014. As a result, the parliament also went through a complex phase of faction evolution, making it an interesting dataset to work with.

Unlike parliamentary voting in countries like the US, the voting in the Ukrainian Parliament is not majorly dictated by party lines. This leads to the formation of non-trivial factions in the Ukrainian parliament. We try to predict how the current set of factions $\vec x_t$ would evolve at a future time step $\vec x_{t+1}$. 

\subsubsection{Dataset Preparation Details}
We use the parliament voting data from \citet{ukraine_data}. It consists of voting and bills introduced from 12th December 2012 (when the new parliament term began) to 6th February 2014 (weeks before the revolution). We construct time steps out of the dataset by clubbing together three consecutive tabled bills to form one time step. 

Our method described in section \ref{sec:methodology} assumes that the factions till time step $t$ are already generated and given to the algorithm as input. Hence, we skip the detailed description of the algorithm generating the factions from the raw voting data as this does not really form the part of our algorithm. However, in summary, following the literature \citep{community_label, ukraine_data}, we construct a network based on the co-voting data of MPs and apply Louvain community detection algorithm \citep{louvain} to get the factions. Finally, we ended up with 164 time steps, which were split into train, test, and validation parts as given in the table \ref{table:data_splits}.

\begin{table}[t]
\begin{center}
\begin{tabular}{|l|l|l|l|}
\hline \bf  & \bf Train & \bf Validation & \bf Test\\ \hline
Ukrainian Parliament & 130 & 10 & 24\\
EU-email & 85 & 5 & 26\\
\hline
\end{tabular}
\end{center}
\caption{\label{table:data_splits} Train, test, and validation split for the two datasets used. The small sizes of the datasets represent one of the main challenges of this domain.}
\end{table}

\subsubsection{Architecture Details}

Sinkhorn networks in their most basic form consist of a neural network followed by the operation of Sinkhorn iterations instead of the usual softmax function. Although Sinkhorn networks are very general and one can use LSTM or CNN-based versions, for this task we go with simple time-lagged networks with Markov assumption of order three. Hence, our input consists of: \,
$f_{\text{inp}} = [\mat[x]_{t}:\vec b_t:\mat[x]_{t-1}:\vec b_{t-1}:\mat[x]_{t-2}]$. Here, $\vec b_t$  represents the exogenous features that capture the historical mass flow. That is, $\vec b_t$ captures the mass flow that had happened between $\vec x_{t-1}$ and $\vec x_{t}$ and $\vec b_{t-1}$ captures the mass flow that had happened between $\vec x_{t-2}$ and $\vec x_{t-1}$. Further, for practical reasons, we limit the number of iterations for the Sinkhorn operator to 100. All the results presented below use the best value hyperparameter $\lambda$ in the equation \ref{eq:objective}. 

\subsubsection{Baseline Algorithms}

We compare our algorithm with a few standard baselines. The baselines can be majorly divided into two categories. In the first category, we have baselines that are more statistically inspired. These combine the previous flow matrices in a few basic ways. 
\begin{enumerate}
    \item {\bf Identity}: assumes that factions at time step $t+1$ would remain exactly the same as factions at time step $t$.
    \item {\bf Average history}: predicts the flow by averaging the flow from the previous two time steps. Note, we are averaging out the previous two flows only in order to match the Markov order 3 assumption of our Sinkhorn-Flow model and hence doing a fair comparison. 
\end{enumerate}

In the second category of baselines, the algorithms try to predict the future faction of each element (MP for the Ukrainian Parliament Dataset) individually. These individual predictions are then combined to form the flow at a given time step $t$.  

\begin{enumerate}
    \item {\bf Logistic Regression (LR)}: a basic logistic regression model. 
    \item {\bf Multi-layer preceptron (MLP)}: a feed-forward neural net model with almost the same architecture as Sinkhorn-Flow. The difference is that instead of predicting a transport plan, a k-class classification is done for predicting the factions at time step $t$. Again, just like LR, predictions are made for each individual element which are then aggregated to form predicted mass flow. 
\end{enumerate}

The input features for both LR and MLP baselines were the previous $m$ faction assignments of the given element, where $m$ is set to three for a fair comparison with Sinkhorn-Flow. 
   
We compare the four baselines mentioned above with our proposed algorithm, Sinkhorn-Flow in table \ref{table:ukrain_overall}. Sinkhorn-Flow was run multiple times with different random seeds and the results averaged to minimize any variations due to weight initialization. 

We compare the different baselines on two metrics: 
\begin{enumerate}
    \item Flow Cost: Frobenious distance between the predicted transport $\hat{\mat[P]}_{t+1}$ and the ground truth transport plan $\mat[P]_{t+1}$. 
    \item Faction Cost: Root Mean Squared Error (RMSE) between the predicted factions and the ground truth factions at time step $t+1$. 
\end{enumerate}

\begin{table}[t]
\begin{center}
\begin{tabular}{|c|c|c|}
\hline \bf  & \bf Flow Cost & \makecell{\bf Faction Cost \\ (RMSE) (1e-2)}\\ \hline
 \hline
Identity baseline & 140.94 & 6.60  \\
Average history & 85.92 & 6.10  \\
Sinkhorn-Flow (our) & \bf 82.22 & \bf 5.96  \\
LR baseline & 136.57 & 8.47  \\
MLP baseline & 119.24 & 8.39  \\
\hline
\end{tabular}
\end{center}
\caption{\label{table:ukrain_overall} Average testing data performance on Ukrainian parliament voting dataset. Flow cost tries to capture the distance between the predicted transport plan and the ground truth transport plan. Faction Cost is the root mean squared error between predicted factions and the ground truth factions.}
\end{table}

\subsection{Dataset 2: EU-email}

We use the "EU email department 4" dataset from \citep{eu-email}. This dataset captures the email communication between members of a research institution in Europe. The changing communication pattern over time gives rise to an implicit problem of evolving factions. The communication over three consecutive dates were clubbed together to form one time step. The details of the dataset are mentioned in table \ref{table:data_splits}. 

The architecture and the hyperparameters used for this dataset are same as those for the problem of Ukrainian Parliament Voting. The input features included comprise of just the auto-regressive features of factions ($\vec x_i$) at time steps $t$, $t-1$, and $t-2$.

Table \ref{table:email} compares the performance of Sinkhorn-Flow with other baselines on the EU-email dataset. 

\begin{table}[t]
\begin{center}
\begin{tabular}{|l|l|l|}
\hline \bf  & \bf Flow Cost & \makecell{\bf Faction Cost \\ (RMSE) (1e-2)}\\ \hline
Identity baseline & 29.44 & 5.18  \\
Average history & 27.5 & 5.64  \\
Sinkhorn-Flow (our) & \bf 25.30 & \bf 4.95  \\
LR baseline & 59.09 & 15.59  \\
MLP baseline & 32.05 & 7.14  \\
\hline
\end{tabular}
\end{center}
\caption{\label{table:email} Average testing data performance on EU  email dataset. }
\end{table}

\subsection{Qualitative Results}

\begin{figure}[t]
\begin{center}
\subfloat[Target Flow]{%
  \includegraphics[clip,width=0.7\columnwidth]{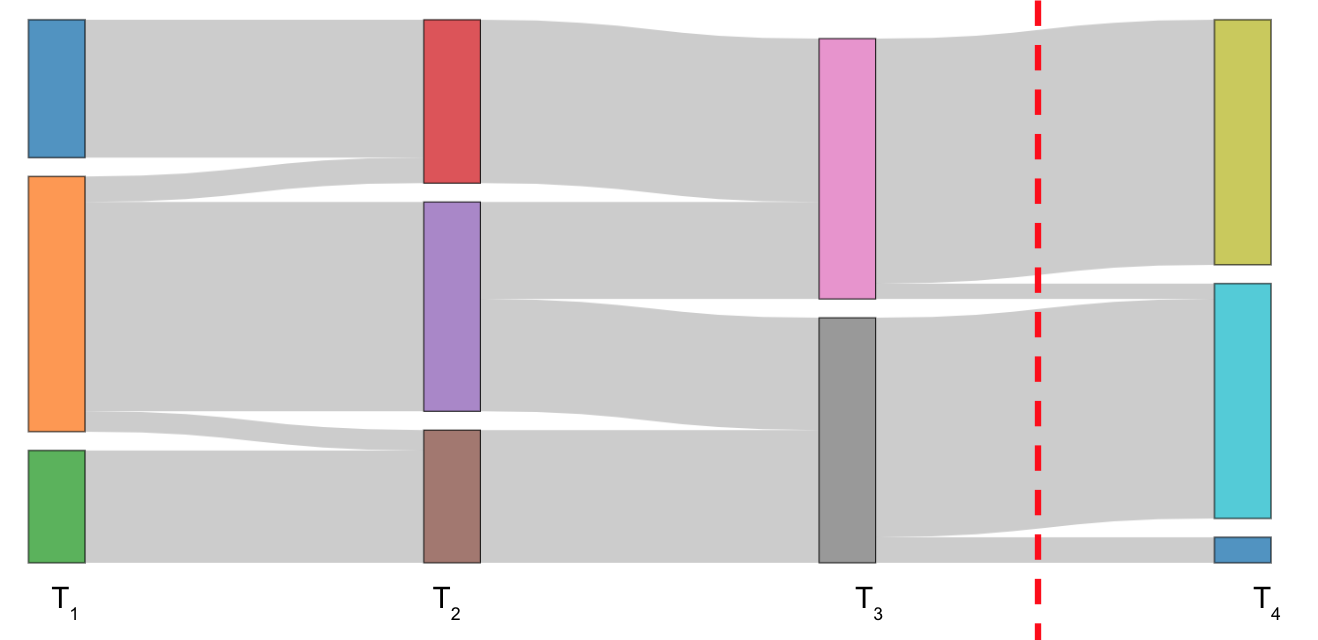}%
}

\subfloat[Predicted Flow]{%
  \includegraphics[clip,width=0.7\columnwidth]{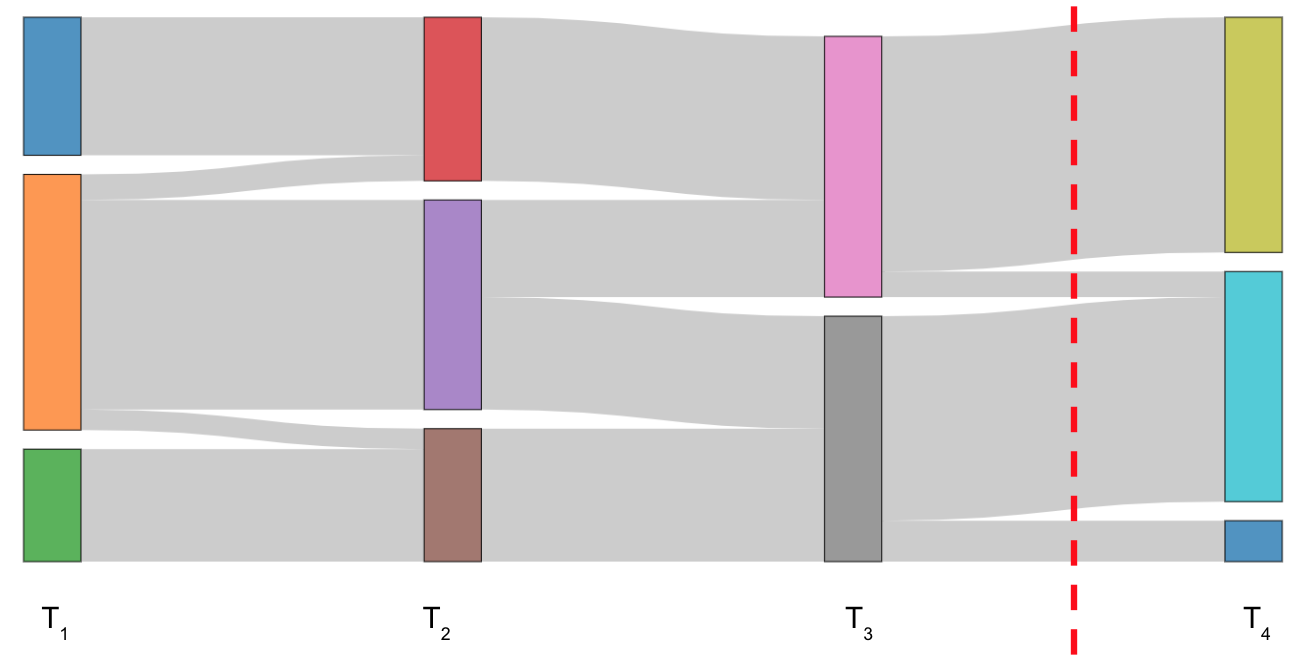}%
}
\end{center}

\caption{The top figure describes the actual evolution of the factions. The time steps to the left of the red-dashed lines are given and the aim is to predict the flow (and factions) to the right of the red line. Thus, in the bottom figure, information to the left of the red line is the input and the factions to the right of the dashed line are the predictions. We can see that our algorithm is able to detect that the two factions at time step $T_3$ would break into 3. Note: the colors do not signify anything.}
\label{fig:eg1}
\end{figure}

\begin{figure}[t]
\begin{center}
\subfloat[Target Flow]{%
  \includegraphics[clip,width=0.7\columnwidth]{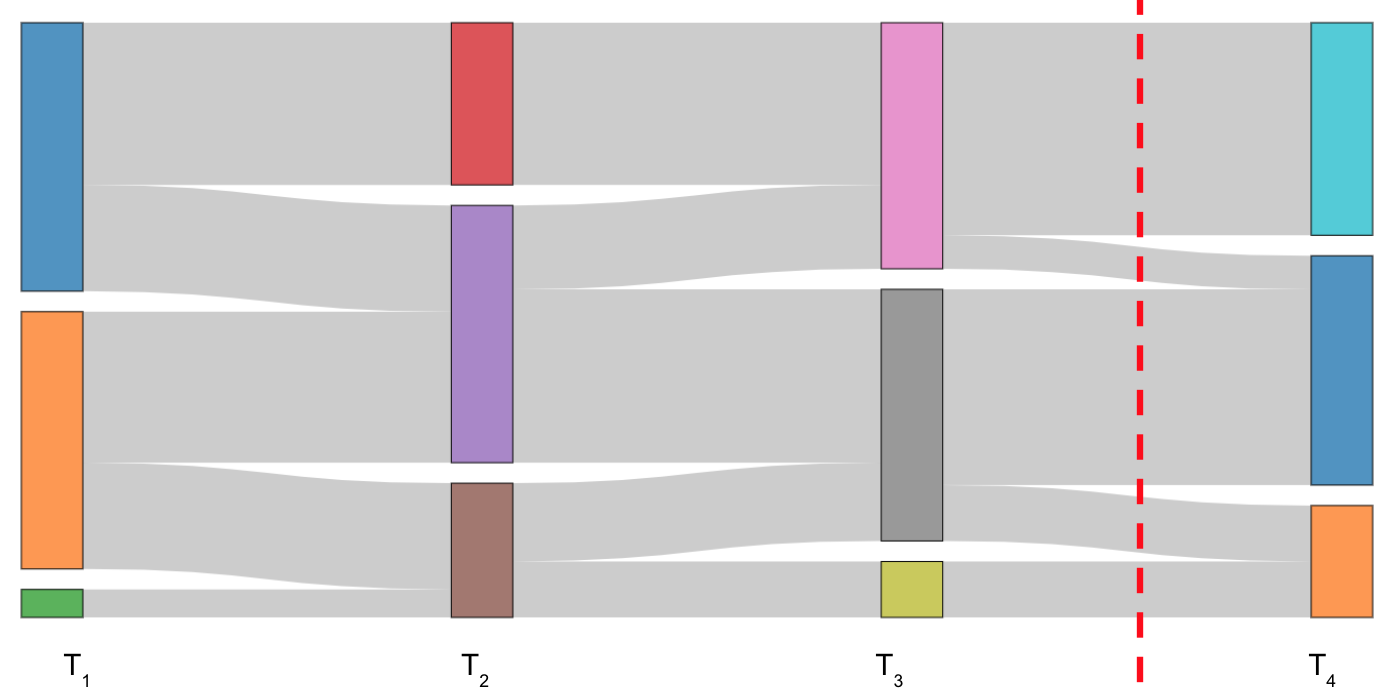}%
}

\subfloat[Predicted Flow]{%
  \includegraphics[clip,width=0.7\columnwidth]{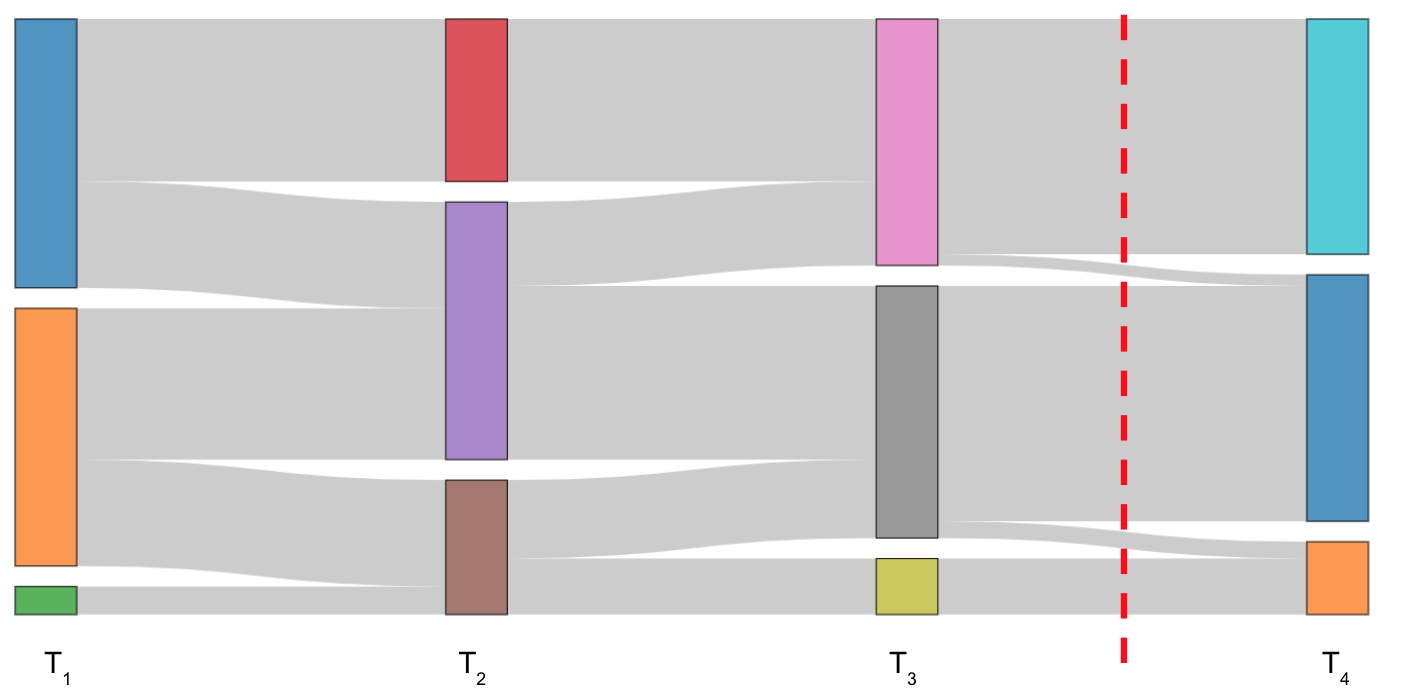}%
}
\end{center}
\caption{(Specifically picked poorly predictive example): Our algorithm is able to predict that the mass of the third faction would increase. Although the predicted size increase falls short of the actual size increase.}
\label{fig:eg3}
\end{figure}

We show a few cherry-picked qualitative results (both good and poor) for the task of predicting the evolution of factions in the Ukrainian parliament in figures \ref{fig:eg1} and \ref{fig:eg3}. These Sankey diagrams show the ground truth flow (top) and the predicted flow of our model (bottom). In both the figures, the first three time steps ($T_1, T_2$ and $T_3$) were given as input ($\vec x_{t-2}, \vec x_{t-1}, \vec x_t$) and the factions at time step $T_4$ as well as the flow between time steps $T_3$ and $T_4$ was predicted.

\subsection{Multi-Step Prediction}

Apart from predicting the flow for the immediate future, a common use case may be of predicting the flow over multiple time steps in the future.

A real-world example would be to predict pseudo-lineage in cells in biological systems \citep{cell_lineage} or predicting how factions evolve over extended periods of time in a social network. 

To show the efficacy of our method for the above use case, we predict $k$ future time steps instead of just one. 
Thus, along with predicting the mass flow between time steps t and t+1, we also predict the flow between t+1 and t+2, t+2 and t+3.... t + (k-1) and t+k. 

One caveat to note here is that all exogenous features (denoted earlier as $\vec b_t$) should either come as a byproduct of the prediction or should be available for future time stamps as well. Both of our current examples fall in the former category as we use the predicted flow at time step t+j as exogenous features to predict flow at time step t + j + 1.

If the above caveat about the exogenous features is satisfied, one can theoretically extend this process to predict flow indefinitely. However, in any auto-regressive setting, any error in prediction is magnified in the future. This practically limits the number of time steps which can be predicted in the future with reasonable accuracy. 

Table, \ref{table:multi_step} describes the results for predicting flows for $k$ time steps into the future. Specifically, it describes the summation of flow cost for predicting ahead for 3 and 5 time steps into the future.

\begin{table}[t]
\begin{center}
\begin{tabular}{|c|c|c|}
 \hline
\multicolumn{3}{|c|}{Ukrainian Parliamentary Voting} \\
 \hline
 & \multicolumn{2}{|c|}{\bf Flow Cost}\\
 \hline
 & k = 3 & k = 5\\ 
 \hline
 Identity baseline  & 451.70 & 749.44\\
 Average history   & 342.04 & 612.15\\
 Sinkhorn-Flow (ours) & \bf 324.90 & \bf 593.72\\
 LR baseline   & 736.92 & 1418.18\\
 MLP baseline   & 410.57 & 728.746\\
 \hline
 \hline
 \multicolumn{3}{|c|}{EU-Email} \\
 \hline
 & k = 3 & k = 5\\
 \hline
 Identity baseline   & 112.34    & 218.24\\
 Average history   & 112.19    & 222.95\\
 Sinkhorn-Flow (ours)   & \bf 106.63    & \bf 167.92\\
 LR baseline   & 175.15    & 233.20\\
 MLP baseline   & 117.01    & 317.33\\
 \hline
\end{tabular}
\end{center}
\caption{\label{table:multi_step} Flow Cost for comparing the predicted flows for three and five time steps into the future. Note that cost is cumulative, i.e, the cost covers the distance between the predicted flow and the ground truth flow for 1 to k time steps. }
\end{table}

\subsection{Discussion}

From tables \ref{table:ukrain_overall} and \ref{table:email}, we can see that Sinkhorn-Flow outperforms all the baselines in both the metrics (flow cost and RMSE) for both the datasets. This shows that our model is able to predict both the flow as well as the marginals better than other techniques that were employed.

Figure \ref{fig:eg1} visually shows an example in which our model is able to predict the flow pretty well. Here, the vertical bars show the marginals and the grey streams show the flows. Although both the ground truth (top) and the predicted (bottom) Sankey diagrams in figure \ref{fig:eg1} look eerily similar, minute differences can be seen upon close inspection. 

Figure \ref{fig:eg3} shows the case in which the difference between ground truth and the predicted flow is more obvious. Although our algorithm does predict that a part of faction 2 at time step $T_3$ would break away and join faction 3 at time step $T_4$, the predicted mass is lesser than the actual mass which got transferred in the ground truth.

Table \ref{table:multi_step} shows that our model performs reasonably well even when we do multi-time step prediction. Although the performances of all the models decrease as the number of time steps to be predicted increases, our model is still able to perform far better than all the other alternatives present.  

Finally, although both the examples presented here did involve an underlying network, we would like to emphasize the fact that our algorithm works for any system which would involve entities exchanging masses over time. Other examples of such problems might be seeing how factions evolve in electoral districts of a nation etc.

\section{Conclusions}

In this paper, we have presented Sinkhorn-Flow, an approach that leverages optimal transport to predict mass flow in a time series forecasting setting.  The technique is based upon past work on Sinkhorn networks \citep{mena2018}, but extends them from an algorithmic standpoint and from an application perspective considers these methods in the specific context of forecasting mass flow in time series. We highlight the application of the method to two real-world applications in community forecasting and show the approach improves substantially over several competitive baseline approaches.

\newpage
\bibliographystyle{named}
\bibliography{ijcai20}

\end{document}